\documentclass{article}
\usepackage{spconf,amsmath,graphicx}
\usepackage{multirow}
\usepackage{hyperref}
\usepackage{subcaption}
\hypersetup{ colorlinks=false, linkcolor=blue, filecolor=black, urlcolor=blue, }

\title{Test-time Adaptation for Out-of-Distributed Image Inpainting}
\name{Chajin Shin\sthanks{E-mail: chajin@yonsei.ac.kr}, Taeoh Kim, Sangjin Lee, Sangyoun Lee\sthanks{Corresponding Author, E-mail: syleee@yonsei.ac.kr}}
\address{School of Electrical and Electronic Engineering, Yonsei University, Seoul, Korea}

\begin{document}
%\ninept
%
\maketitle
\begin{abstract}
	Deep-learning-based image inpainting algorithms have shown great performance via powerful learned priors from numerous external natural images. 
	However, they show unpleasant results for test images whose distributions are far from those of the training images because their models are biased toward the training images.
	In this paper, we propose a simple image inpainting algorithm with test-time adaptation named \textit{AdaFill}.
	Given a single out-of-distributed test image, our goal is to complete hole region more naturally than the pre-trained inpainting models.
	To achieve this goal, we treat the remaining valid regions of the test image as an another training cue because natural images have strong internal similarities.
	From this test-time adaptation, our network can exploit externally learned image priors from the pre-trained features as well as the internal priors of the test image explicitly.
	The experimental results show that \textit{AdaFill} outperforms other models on various out-of-distribution test images.
	Furthermore, the model named \textit{ZeroFill}, which is not pre-trained also outperforms the pre-trained models sometimes.
	
\end{abstract}

\section{Introduction}
\label{sec:intro}
Image inpainting is a task that complete a missing region in an image using information from the valid regions. 
We can remove unwanted objects, text, or scratches via image inpainting techniques.
Following rapid development of deep-learning-based imaging algorithms, most of the image inpainting models are trained with extensive training images.
Some methods have been proposed to handle arbitrary mask shapes~\cite{GatedConv, PartialConv} or to capture semantic structures using novel architectures or loss functions~\cite{EC, CA}.
These models can learn good natural image priors from huge dataset and are good at processing textures and structures that frequently appear in the training dataset.

\begin{figure}[t]
	\captionsetup[subfigure]{labelformat=empty}
	\centering
	\begin{subfigure}[c]{0.24\linewidth}
		\centering
		\includegraphics[width=\linewidth]{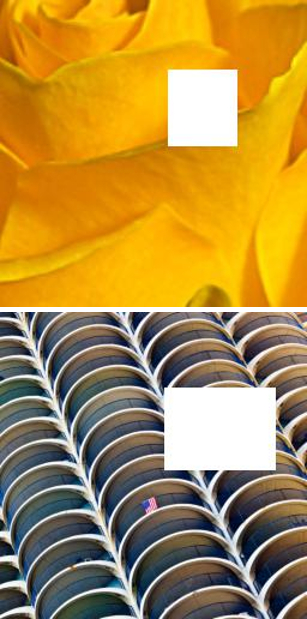}
		\caption{\footnotesize{Input}}
	\end{subfigure}
	\begin{subfigure}[c]{.24\linewidth}
		\centering
		\includegraphics[width=\linewidth]{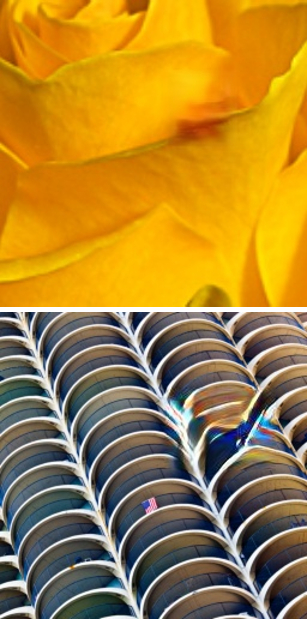}
		\caption{\footnotesize{GatedConv \cite{GatedConv}}}
	\end{subfigure}
	\begin{subfigure}[c]{.24\linewidth}
		\centering
		\includegraphics[width=\linewidth]{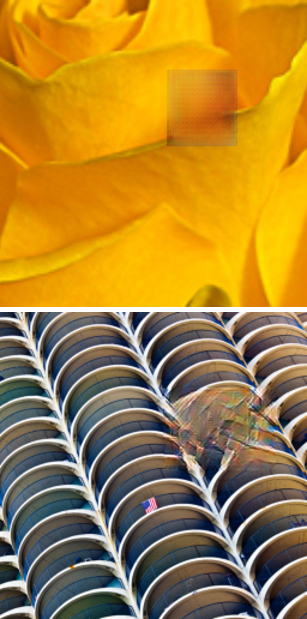}
		\caption{\footnotesize{EdgeConnect \cite{EC}}}
	\end{subfigure}
	\begin{subfigure}[c]{.24\linewidth}
		\centering
		\includegraphics[width=\linewidth]{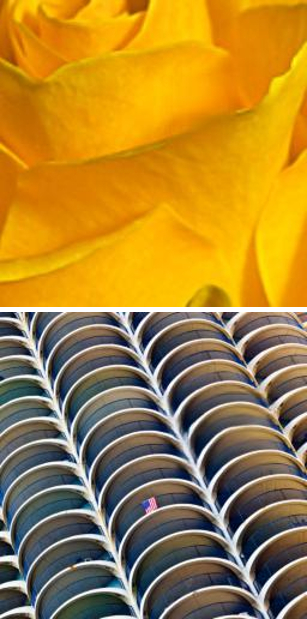}
		\caption{\footnotesize{\textbf{AdaFill(Ours)}}}
	\end{subfigure}
	\caption{GatedConv~\cite{GatedConv} and EdgeConnect~\cite{EC} show splashing or diffusing artifacts and cannot grasp internal similarity. In contrast, our method recover with less artifacts. }
	\label{fig:introduction}
\end{figure}

\begin{figure*}[!ht]
	\centering
	\begin{subfigure}[c]{1.0\textwidth}
		\centering
		\includegraphics[width=1.0\textwidth]{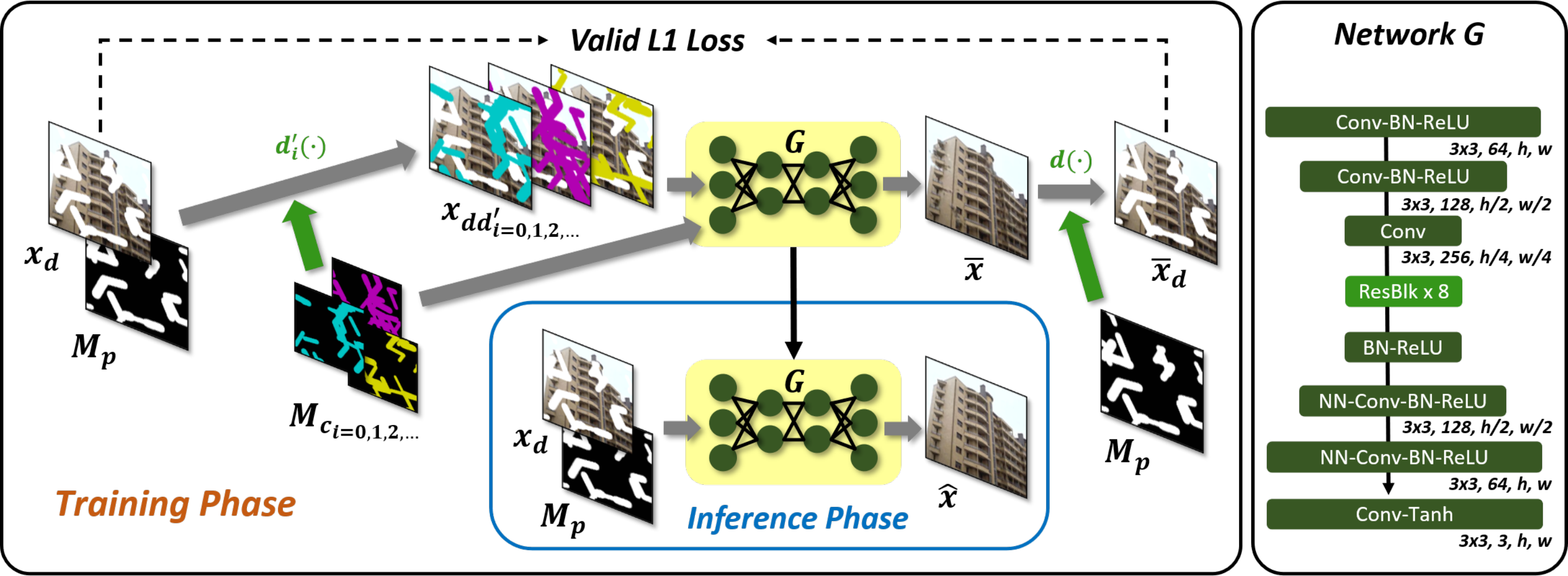}
	\end{subfigure}
	
	\caption{\textbf{Left}: Overall flow. In the training phase, we begin from the pre-trained inpainting network $G$ (or random initialization for \textit{ZeroFill}). Next, for the test-time adaptation, we degrade a test image $x_d$ with random child masks $M_{c_{i=0, 1, 2, ...}}$ and put them into the network. The output of the network has to be same with the test image $x_d$ at the valid regions. After test-time training, the test image $x_d$ is passed along with its parent mask $M_p$ to get the final inpainted image. \textbf{Right}: Structure of the inpainting network $G$.}
	\label{fig:overall scheme}
\end{figure*}

However, these models have difficulty in recovering images whose patterns are totally different from those of the training images.
This domain gap causes severe color artifacts, and these models cannot exploit internally similar patches that appear in the test images, as depicted in Fig.~\ref{fig:introduction}. 
As shown in Fig.~\ref{fig:introduction}, the color patterns are not consistent with the surrounding regions, which causes artifacts. 

To cope with this problem, inspired by the recent internal learning algorithms~\cite{DIP, ZeroShotSR}, we propose a test-time adaptation algorithm for image inpainting named \textit{AdaFill}. 
First, we modify the existing inpainting networks to fit a single test image and pre-train the network on a large-scale dataset to achieve external image priors. Next, we train the model on the test image only so that the network can focus on the internal pixel distribution by exploiting valid regions explicitly.
With this simple scheme, our model can handle color artifacts caused by domain gap and exploit the internal similarity of a test image. 
We also propose a non-pre-trained version of \textit{AdaFill}, called \textit{ZeroFill}, that shows performance comparable with the pre-trained models.
To the best of our knowledge, this is the first work that tackles the distributional shift problem in image inpainting.
Compared with the other restoration tasks, the restoration performance in image inpainting is more dependent on the training dataset. 
Therefore, generalization for out-of-distributed images is an important issue for practical usage.

\section{Related Works}
\label{sec:preliminary section}

Natural images have a high unique internal similarity, where similar structures or textures across various scales appear recurrently within an image. 
Several studies have previously verified that such internal similarity can be utilized for the single image super-resolution task~\cite{ZeroShotSR, zontak2011internal, glasner2009super}. 
They show that the internal statistical prior from a single image is powerful and often better than the generalized statistics from the large-scale training.

Our work is closely related to the ZSSR~\cite{ZeroShotSR} that perform image super-resolution from a single image via internal learning. They artificially generate training samples from a low-resolution test image using re-downsampling. 
Compared with the ZSSR that exploits whole degraded images, our method utilizes valid regions as strong training cues.
Similarly, DIP~\cite{DIP} propose a method that implicitly learns the prior of a single image. 
They show that this internal prior can be used to recover images with various types of degradations including image inpainting. However, such implicit internal prior has difficulty in recovering extreme degradations such as large holes or holes with extreme non-local patterns.
In contrast, our method explicitly learns the internal prior using artificial training samples as well as external prior using large-scale pre-training.

\begin{table*}[t!]
	\centering
	\resizebox{\textwidth}{!}{%
		\begin{tabular}{c|ccc|ccc|ccc|ccc}
			Model & \multicolumn{3}{c|}{Gated Conv~\cite{GatedConv}} & \multicolumn{3}{c|}{Edge Connect~\cite{EC}} & \multicolumn{3}{c|}{DIP~\cite{DIP}} & \multicolumn{3}{c}{AdaFill(Ours)} \\ \hline
			Dataset & PSNR & SSIM & LPIPS & PSNR & SSIM & LPIPS & PSNR & SSIM & LPIPS & PSNR & SSIM & LPIPS \\ \hline\hline
			T91~\cite{OtherDataset} & 27.15 & 0.889 & 0.0755 & \textbf{27.85} & \textbf{0.901} & \textbf{0.0692} & 25.46 & 0.851 & 0.1047 & 27.26 & 0.870 & 0.0843 \\
			Urban100~\cite{Urban100} & 23.14 & 0.854 & 0.0722 & 24.06 & \textbf{0.866} & 0.0721 & 22.23 & 0.805 & 0.1082 & \textbf{24.52} & 0.845 & \textbf{0.0699} \\
			Google map~\cite{GoogleMap} & 24.65 & 0.846 & 0.0939 & 26.25 & 0.848 & \textbf{0.0888} & 24.47 & 0.836 & 0.1164 & \textbf{26.73} & \textbf{0.858} & 0.0910 \\
			Facade~\cite{Facade} & 26.25 & 0.900 & 0.0570 & 26.02 & 0.886 & 0.0688 & 25.79 & 0.890 & 0.0691 & \textbf{28.55} & \textbf{0.921} & \textbf{0.0451} \\
			BCCD~\cite{BCCD} & 34.25 & 0.956 & \textbf{0.0595} & \textbf{34.43} & 0.954 & 0.0662 & 30.47 & 0.948 & 0.0890 & 34.26 & \textbf{0.962} & 0.0631 \\
			KLH~\cite{KLH} & 33.25 & \textbf{0.823} & \textbf{0.1162} & 20.91 & 0.791 & 0.3419 & 30.33 & 0.751 & 0.1579 & \textbf{33.48} & 0.781 & 0.1919 \\
			Document~\cite{Document} & 19.67 & 0.910 & 0.0762 & 18.84 & 0.876 & 0.1316 & 17.49 & 0.865 & 0.1122 & \textbf{20.72} & \textbf{0.919} & \textbf{0.0585}
		\end{tabular}%
	}
	\vspace{-0.1cm}
	\caption{Quantitative comparison with GatedConv~\cite{GatedConv}, EdgeConnect~\cite{EC}, and DIP~\cite{DIP}}
	\vspace{-0.4cm}
	\label{tab:model comparison}
\end{table*}

\section{Method}
\label{sec:method}
Our overall framework and network structure are described in Fig.~\ref{fig:overall scheme}.
In the training phase, we begin from the pre-trained inpainting network $G$ and we fine-tune on a single degraded image $x_d$. For \textit{ZeroFill}, we skip the pre-training process.
We assume that the image $x_d$ is distorted by a distortion function $d(\cdot)$ with a parent mask $M_p$ from the clean image $x$ like $x_d=d(x)$. 
The parent mask $M_p$ represents the invalid pixels of $x_d$ with value 1 and valid pixels of $x_d$ with value 0. 
Therefore, we can represent the distorted image as $ x_d = x \odot (1 - M_p) + M_p $,
where $\odot$ represents element-wise multiplication. 
To enable our network to learn the internal similarity and exploit it for inpainting, we define a similar distortion function $d'(\cdot)$ with the child mask $M_c$. 
As illustrated in Fig.~\ref{fig:overall scheme}, we degrade the given image using the child mask $d'(x_d) = x_d \odot (1 - M_c) + M_c$.
The child masks are randomly generated during training. 
We denote this double-distorted image as $x_{dd'}$. 
In the case where the shape of the parent mask $M_p$ is different from the irregular mask or box mask, we use the parent mask as a child mask with random rotation and scaling at a certain rate.

Our goal is to make the network learn the mapping function from the double-distorted image $x_{dd'}$ to the given single distorted image $x_d$. 
\begin{equation}
	\bar{x} = G([x_{dd'}, M_c])
	\label{equation:eq2}
\end{equation}

\noindent where $\bar{x}$ is a preliminary prediction, $G$ is our inpainting network, and $[\cdot, \cdot]$ is concatenation operation.
The predicted image $\bar{x}$ has to be same with the given image $x_d$ for the valid pixels. Since we do not know the ground truth of the invalid pixels in $x_d$, we degrade $\bar{x}$ with the same distortion function $\bar{x}_d = d(\bar{x})$.
After degradation, we use following $L1$ loss to train the inpainting network $G$.
\begin{equation}
	\mathcal{L}_{L1} = |\bar{x}_d - x_d|_1
	\label{equation:eq4}
\end{equation}

From this training step, the network $G$ can learn the restoration patterns in the degraded image using the valid regions while ignoring the parent distortions.

At the inference phase, we proceed one forward propagation with the test image that was used for training with equation, $\hat{x} = G([x_d, M_p])$.  Here, $\hat{x}$ is our final result image. The structure of our inpainting network $G$ is described on the right side of Fig.~\ref{fig:overall scheme}.

%\begin{equation}
%    \hat{x} = G([x_d, M_p])
%    \label{equation:eq4}
%\end{equation}

\noindent

\vspace{-0.5cm}

\section{Experiments}
\label{sec:experiment}

\noindent \textbf{Settings.} We pre-train the network using Places365~\cite{Places365} dataset for 1 epochs with the settings in~\cite{EC}.
For the test-time adaptation, we use the following hyper-parameters: batch size 8, learning rate 0.0001, Adam~\cite{Adam} optimizer with $\beta_1 = 0.5, \beta_2 = 0.9$, and 1,000 training iterations. For \textit{ZeroFill}, we use 5,000 iterations without pre-training.
We evaluate our model with LPIPS~\cite{LPIPS}, SSIM, and PSNR. 
We compare our model with other models that learn only the explicit prior only (pre-trained models, GatedConv~\cite{GatedConv} and EdgeConnect~\cite{EC}), and only the internal prior only (DIP~\cite{DIP}).

\noindent \textbf{Dataset.} We use various datasets for evaluating our model: T91~\cite{OtherDataset}, Urban100~\cite{Urban100}, Google Map~\cite{GoogleMap}, Facade~\cite{Facade}, BCCD~\cite{BCCD}, KLH~\cite{KLH}, BSD200~\cite{BSD}, and Document~\cite{Document}.
These are out-of-distributed from the Places365~\cite{Places365} dataset whose distribution is focused on various places images. In contrast, these images are small objects, natural scenes, artificial structures, medical images, satellite images, or text images.
We subsample and pre-process each dataset and use two type of holes: box mask and irregular mask. For detailed description, please refer to the supplementary materials.

\begin{figure}[t]
	\begin{minipage}[b]{1.0\columnwidth}
		\centering
		\centerline{\includegraphics[width=1.0\textwidth]{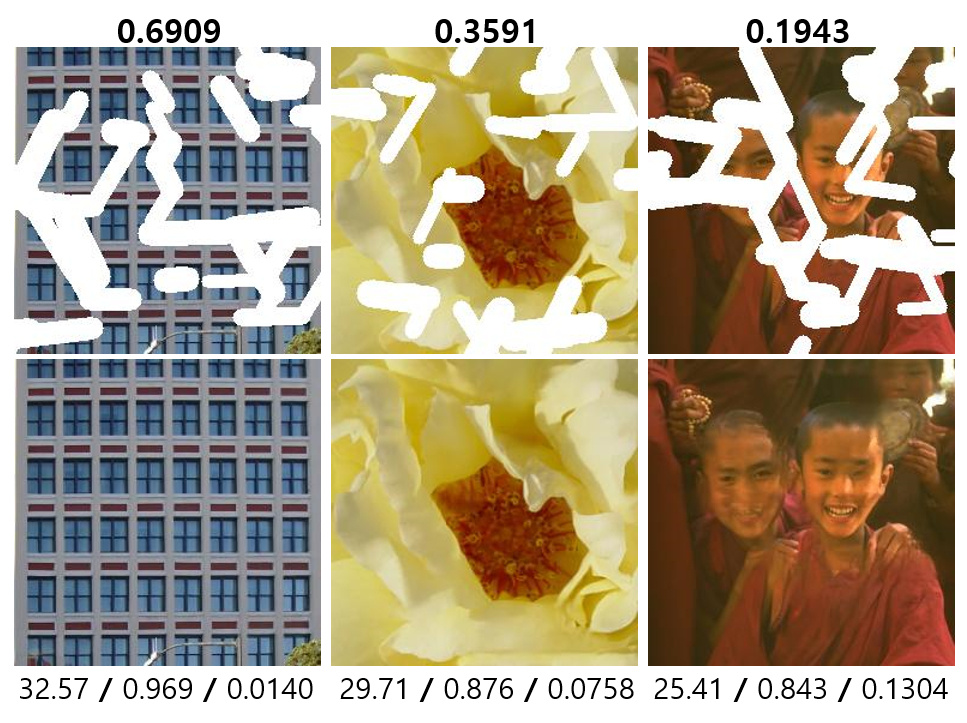}}
	\end{minipage}
	\caption{Values above the images represent the internal similarity scores, and values below the images indicate the PSNR / SSIM / LPIPS~\cite{LPIPS}. The results show that the higher internal similarity scores leads to the better results with our model.}
	\label{fig:internal similarity}
\end{figure}

\subsection{Experimental Results}
\label{ssec:quantitative result}

Quantitative results are described in Table~\ref{tab:model comparison}. From these results, our model outperforms DIP for all datasets and metrics. In almost cases, our model is superior to the pre-trained models even though our model is a one-stage network.
It reveals that exploiting the internal statistics of a test image is critical to image inpainting.
If the dataset has strong internal similarities, such as Urban100, Google Map, and Facade, our model consistently performs better. In addition, if the distribution of the dataset is far from that of the training dataset, such as KLH and Document, pre-trained models cannot recover well.

\begin{figure*}[hbt!]
	\centering
	\rotatebox{90}{\hspace{-5.4cm}\footnotesize \textbf{AdaFill(Ours)}   \quad\quad\quad\;\;    DIP~\cite{DIP}   \quad\quad\quad EdgeConnect~\cite{EC}  \quad\quad  GatedConv~\cite{GatedConv}  \quad\quad\quad\quad\;  Input}
	\begin{subfigure}[c]{0.135\textwidth}
		\centering
		\includegraphics[width=\textwidth]{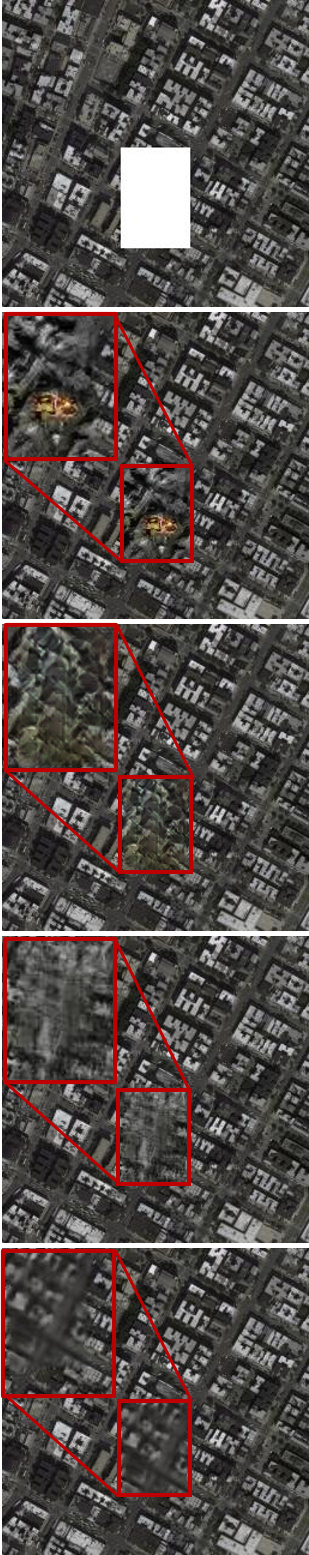}
		\caption*{\footnotesize{Google Map~\cite{GoogleMap}}}
	\end{subfigure}
	\hspace{-0.15cm}
	\begin{subfigure}[c]{.135\textwidth}
		\centering
		\includegraphics[width=\textwidth]{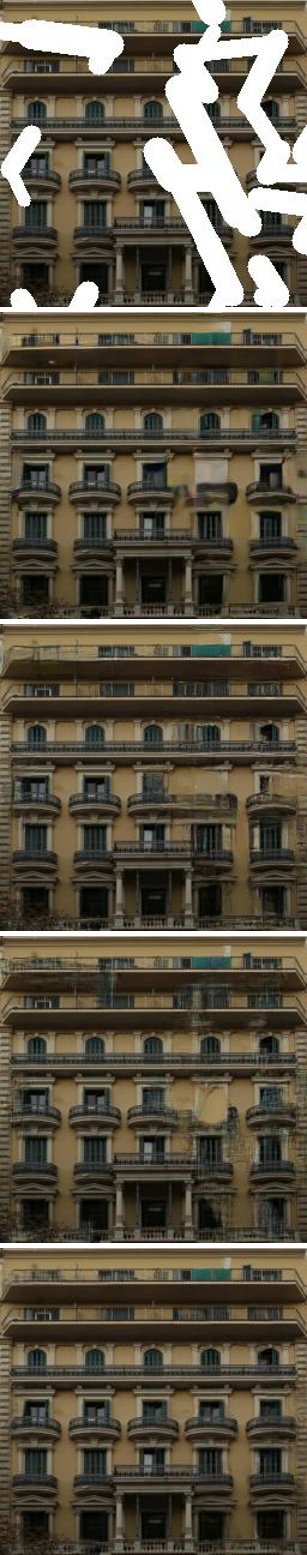}
		\caption*{\footnotesize{Facade~\cite{Facade}}}
	\end{subfigure}
	\hspace{-0.15cm}
	\begin{subfigure}[c]{.135\textwidth}
		\centering
		\includegraphics[width=\textwidth]{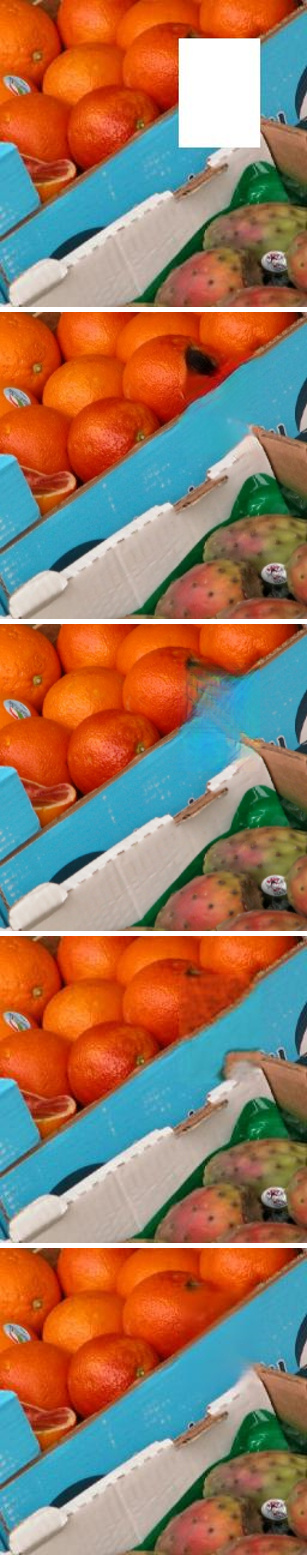}
		\caption*{\footnotesize{T91~\cite{OtherDataset}}}
	\end{subfigure}
	\hspace{-0.15cm}
	\begin{subfigure}[c]{.135\textwidth}
		\centering
		\includegraphics[width=\textwidth]{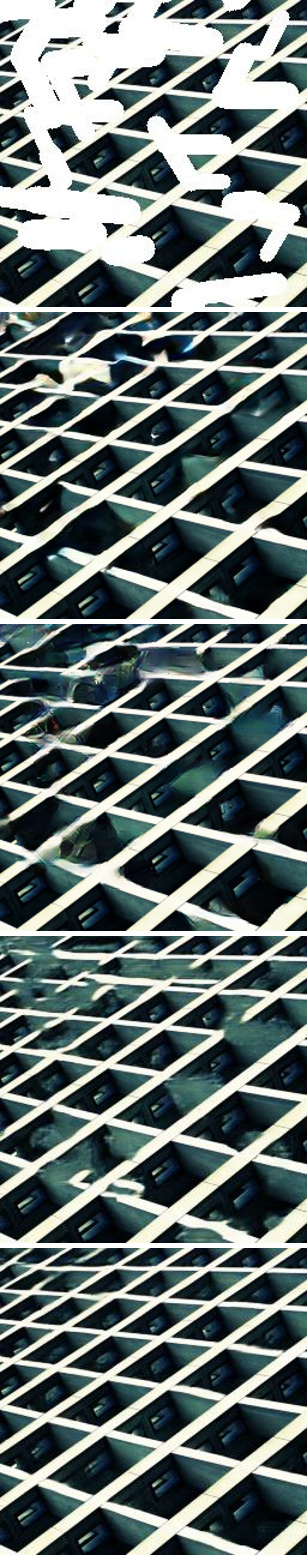}
		\caption*{\footnotesize{Urban100~\cite{Urban100}}}
	\end{subfigure}
	\hspace{-0.15cm}
	\begin{subfigure}[c]{.135\textwidth}
		\centering
		\includegraphics[width=\textwidth]{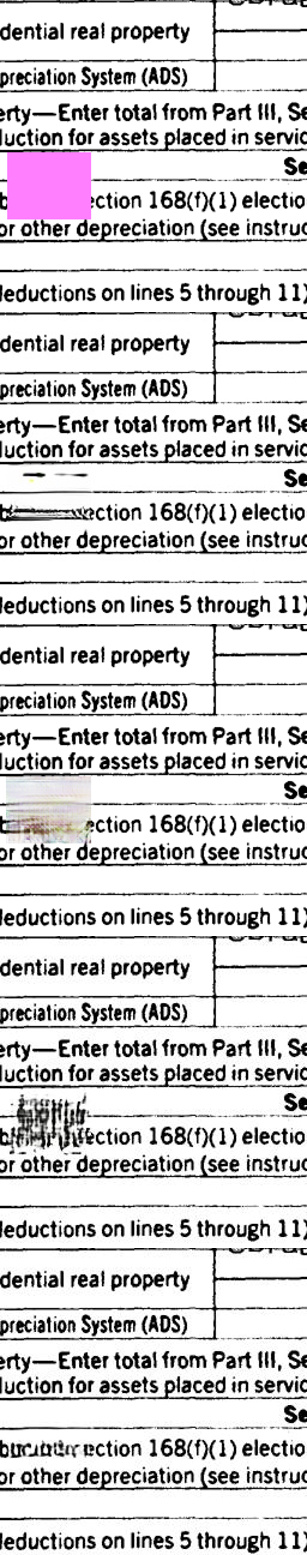}
		\caption*{\footnotesize{Document~\cite{Document}}}
	\end{subfigure}
	\hspace{-0.15cm}
	\begin{subfigure}[c]{.135\textwidth}
		\centering
		\includegraphics[width=\textwidth]{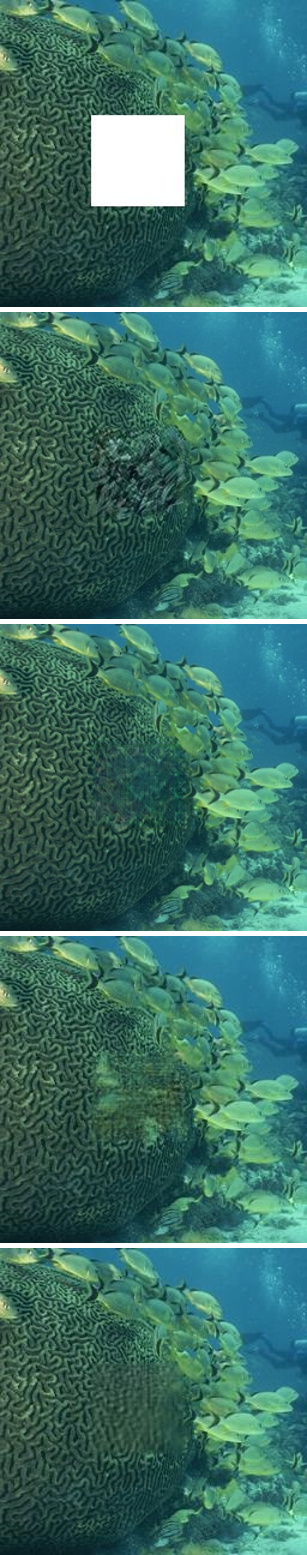}
		\caption*{\footnotesize{BSD200~\cite{BSD}}}
	\end{subfigure}
	\hspace{-0.15cm}
	\begin{subfigure}[c]{.135\textwidth}
		\centering
		\includegraphics[width=\textwidth]{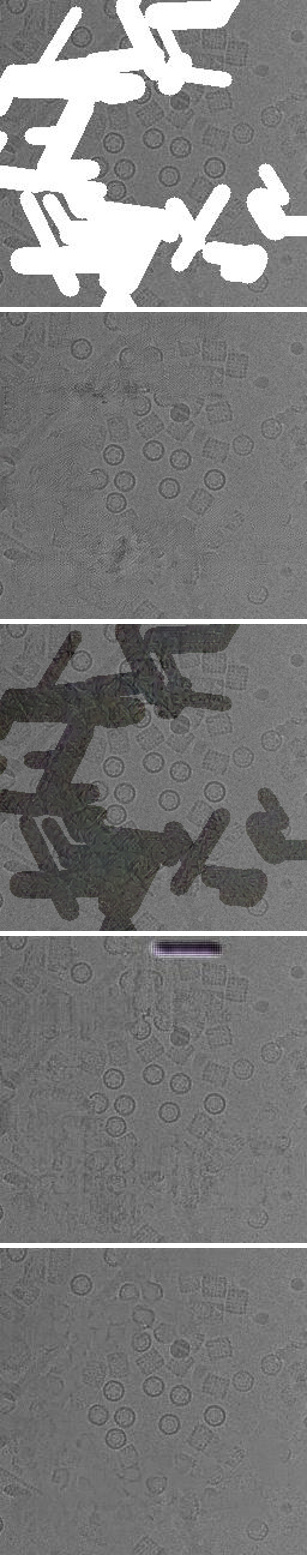}
		\caption*{\footnotesize{KLH~\cite{KLH}}}
	\end{subfigure} 
	\caption{Qualitative comparison results with pre-trained model of GatedConv~\cite{GatedConv}, EdgeConnect~\cite{EC} and DIP~\cite{DIP}. We can confirm that pre-trained models show color artifacts and lower ability in capturing internal similarity in a single image.}
	\label{fig:compare}
\end{figure*}

Our qualitative results are compared in Fig.~\ref{fig:compare}. As mentioned above, similar results are observed. 
In the case of large internal similarity within an image, our model perfectly recover the hole regions, while other models show severe artifacts.

\subsection{Internal Similarity}
\label{sec:internal similiarty}

From the Fig.~\ref{fig:internal similarity}, result show that the higher internal similarity score, the better restoration performance is achieved from our method. 
Our method perfectly recovers the hole regions when the internal similarity is high (the first column of Fig.~\ref{fig:internal similarity}.)
To get the internal similarity, we first use pre-trained VGG19~\cite{VGG} and extract features from \texttt{relu 5-1} layer. Next, we calculate the pixel-wise similarities using the cosine similarity to get the similarity map of size $HW \times HW$, where $H$ and $W$ are the height and width of the feature map, respectively. Finally, we average the similarity map to get the final internal similarity score.

\begin{table}[t!]
	\centering
	\resizebox{\linewidth}{!}{%
		\begin{tabular}{c|cc|cc|ccc}
			& PT      & TTA     & One St.  & BN, NN  & PSNR           & SSIM            & LPIPS           \\ \hline\hline
			EC & $\surd$ &         &         &         & 25.63          & 0.847           & 0.1283          \\
			EC-TTA & $\surd$ & $\surd$ &         &         & 28.52          & \textbf{0.884}  & 0.1095          \\
			EC-TTA & $\surd$ & $\surd$ & $\surd$ &         & 28.57          & 0.883           & 0.0882          \\
			AdaFill & $\surd$ & $\surd$ & $\surd$ & $\surd$ & \textbf{28.57} & 0.882           & \textbf{0.0837} \\
			ZeroFill &        & $\surd$ & $\surd$ & $\surd$ & 27.47          & 0.878           & 0.1108              
		\end{tabular}%
	}
	\caption{Ablation study. PT: pre-training, TTA: test-time adaptation, One St: one-stage network, BN: batch normalization, NN: nearest-neighbor upsampling with convolution.}
	\label{tab:ablation study}
\end{table}

\subsection{Ablation Study}
\label{sec:ablation study}

We conducted ablation studies to find the optimal structure for the test-time adaptation with a single image. 
These results are compared in Table~\ref{tab:ablation study}.
For the ablation experiments, we use the first 10 images from each dataset in Table~\ref{tab:model comparison}.
We modified two things from the EdgeConnect~\cite{EC} baseline structure.
The first one is using only the second stage of the EdgeConnect to reduce the number of parameters.
The second one is replacing instance normalization~\cite{InstanceNorm} $+$ transposed convolution with batch normalization~\cite{BatchNorm} $+$ nearest-neighbor upsampling with convolution.
These modifications increase the perceptual restoration quality, and reduce the color and annoying artifacts a lot.
The results also show that our non-pre-trained model, \textit{ZeroFill} even show a slightly better performance than the pre-trained model.

\section{Conclusion}
\label{sec:conclution}
We propose a simple test-time adaptation scheme called \textit{AdaFill} for image inpainting and \textit{ZeroFill} as an unsupervised version.
The results show that the previous pre-trained models cannot generalize well on the out-of-distributed images.
In contrast, our methods can overcome this domain gap and fully exploit the internal similarity of test images.
As a future works, to reduce the test time, exploiting meta-learning~\cite{MZSR} can be adapted for practical usage.

\vspace*{\fill}
\noindent\footnotesize\textbf{Acknowledgement. }This research was supported by R\&D program for Advanced Integrated-intelligence for Identification (AIID) through the National Research Foundation of KOREA(NRF) funded by Ministry of Science and ICT (NRF-2018M3E3A1057289).
\normalsize

\vfill
\pagebreak

\bibliographystyle{IEEEbib}
\bibliography{strings,refs}

\pagebreak
\onecolumn
	\maketitle

	%\ninept
	%
	
	%
	\section{Supplementary Materials}
	
	\subsection{Experimental Configuration}
	\textbf{Dataset:} For larger than $256 \times 256$ images, we resize images to 256 pixels for shorter axis and random crop to make final resolution of $256 \times 256$. We only use first 100 images for dataset which is set of more than 100 images. In case of Document~\cite{Document} dataset, we just downsample by a factor of 4 and random crop to get the resolution of $256 \times 256$.
	
	\noindent
	\textbf{Mask:} We use irregular mask~\cite{GatedConv} with rate of 10 to 30\%. In case of random box mask, we use 5 to 15\% rate. We average results from irregular and random box mask to get final result value.

	\subsection{Results and Comparisons}
	\subsubsection{BSD200~\cite{BSD} Dataset}
	\begin{figure}[h!]
		\centering
		\begin{minipage}[b]{0.95\columnwidth}
			\centering
			\centerline{\includegraphics[width=1.0\textwidth]{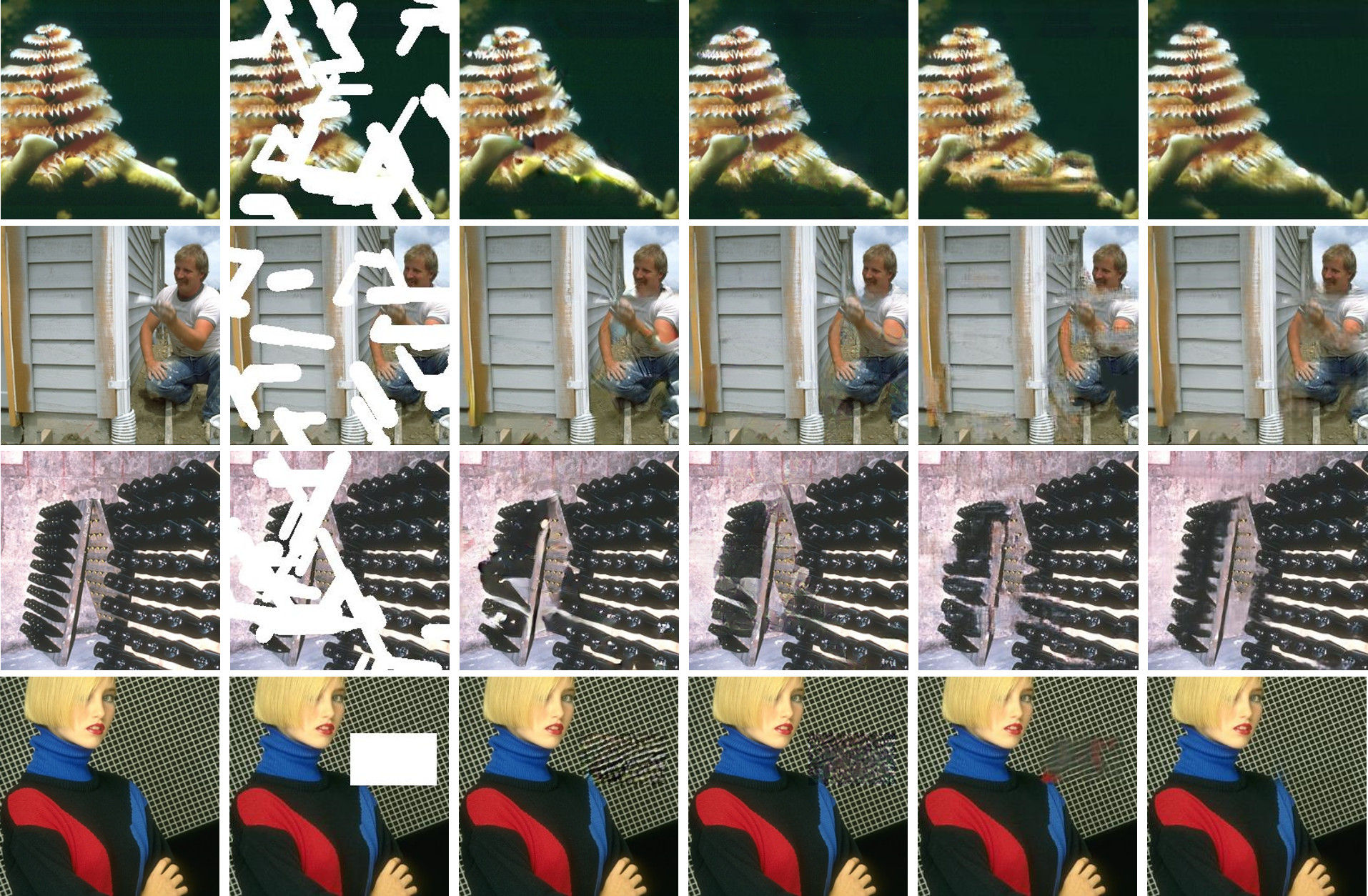}}
			\centerline{GT  \hspace{2cm}   Masked \hspace{2cm}   GC~\cite{GatedConv}  \hspace{2cm}  EC~\cite{EC} \hspace{2cm}  DIP~\cite{DIP} \hspace{2cm}   Ours}
		\end{minipage}
		\caption{Qualitative comparison results with BSD200\cite{BSD} dataset}
		\label{fig:BSD200}
	\end{figure}

	\newpage
	\subsubsection{General100~\cite{OtherDataset} Dataset}
	\begin{figure}[h]
		\centering
		\begin{minipage}[b]{0.95\columnwidth}
			\centering
			\centerline{\includegraphics[width=1.0\textwidth]{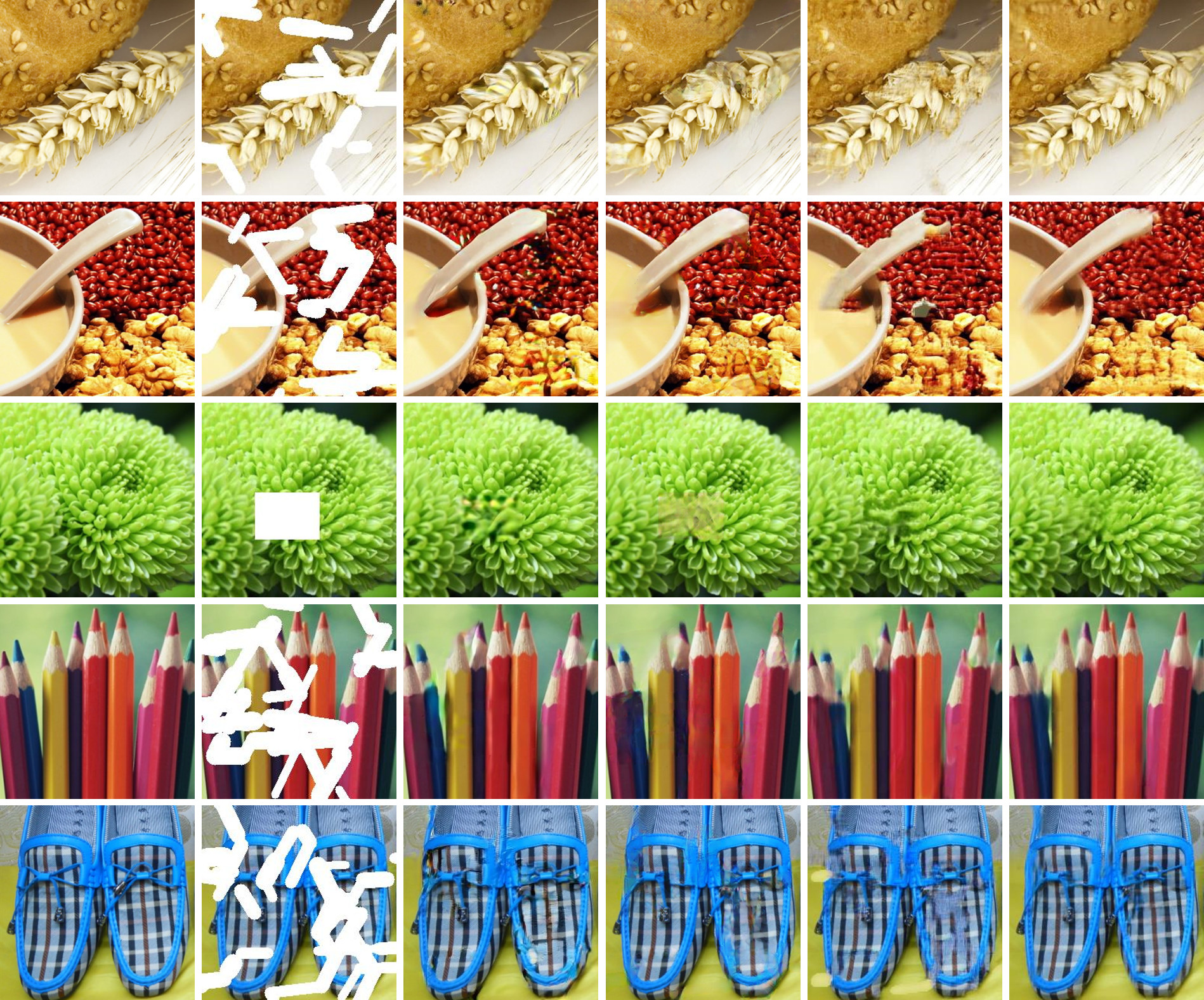}}
			\centerline{GT  \hspace{1.7cm}   Masked \hspace{1.7cm}   GC~\cite{GatedConv}  \hspace{1.7cm}  EC~\cite{EC} \hspace{1.7cm}  DIP~\cite{DIP} \hspace{1.7cm}   Ours}
		\end{minipage}
		\caption{Qualitative comparison results with General100~\cite{OtherDataset} dataset}
		\label{fig:General100}
	\end{figure}

	\subsubsection{T91~\cite{OtherDataset} Dataset}
	\begin{figure}[h!]
		\centering
		\begin{minipage}[b]{0.95\columnwidth}
			\centering
			\centerline{\includegraphics[width=1.0\textwidth]{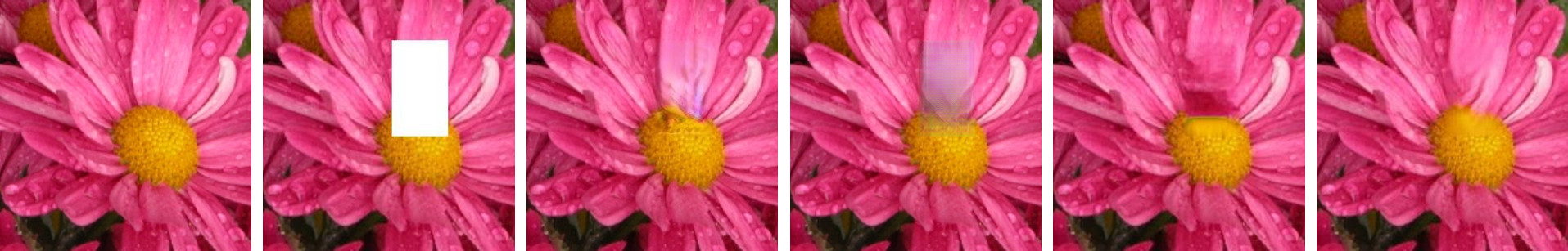}}
			\centerline{GT  \hspace{1.7cm}   Masked \hspace{1.7cm}   GC~\cite{GatedConv}  \hspace{1.7cm}  EC~\cite{EC} \hspace{1.7cm}  DIP~\cite{DIP} \hspace{1.7cm}   Ours}
		\end{minipage}
		\caption{Qualitative comparison results with T91~\cite{OtherDataset} dataset}
		\label{fig:T91-1}
	\end{figure}

	\begin{figure}[h]
		\begin{minipage}[b]{1.0\columnwidth}
			\centering
			\centerline{\includegraphics[width=1.0\textwidth]{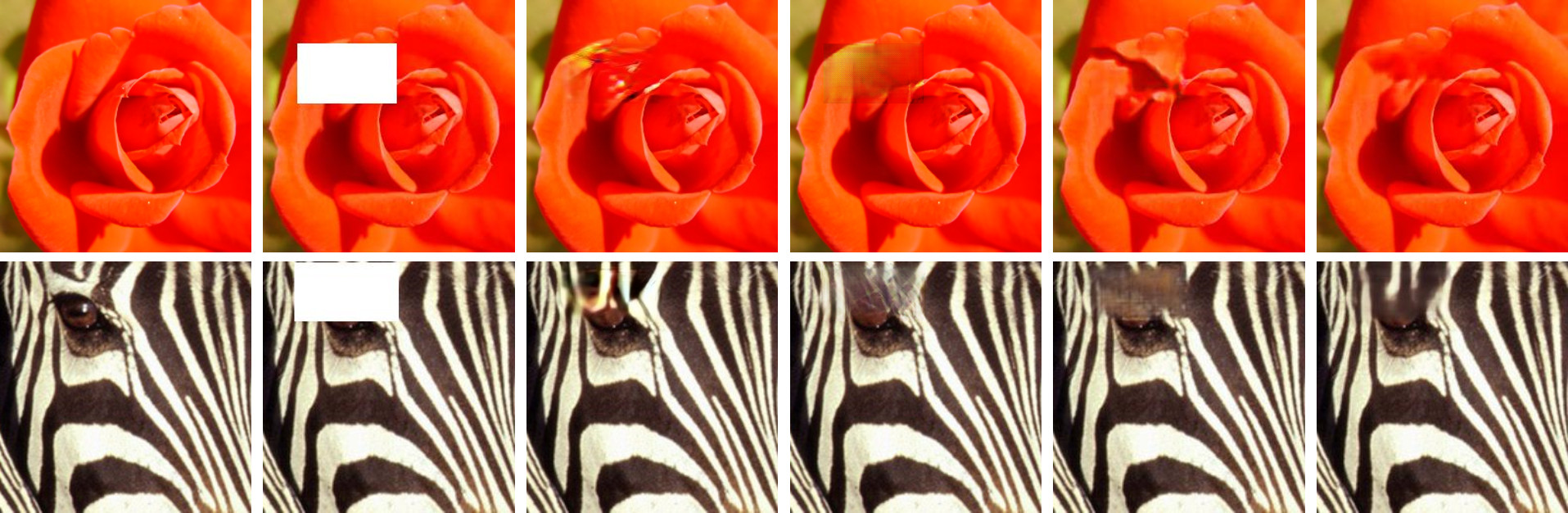}}
			\centerline{GT  \hspace{1.7cm}   Masked \hspace{1.7cm}   GC~\cite{GatedConv}  \hspace{1.7cm}  EC~\cite{EC} \hspace{1.7cm}  DIP~\cite{DIP} \hspace{1.7cm}   Ours}
		\end{minipage}
		\caption{Qualitative comparison results with T91~\cite{OtherDataset} dataset}
		\label{fig:T91-2}
	\end{figure}

	\newpage
	\subsubsection{Urban100~\cite{Urban100} Dataset}
	\begin{figure}[h!]
		\centering
		\begin{minipage}[b]{0.95\columnwidth}
			\centering
			\centerline{\includegraphics[width=1.0\textwidth]{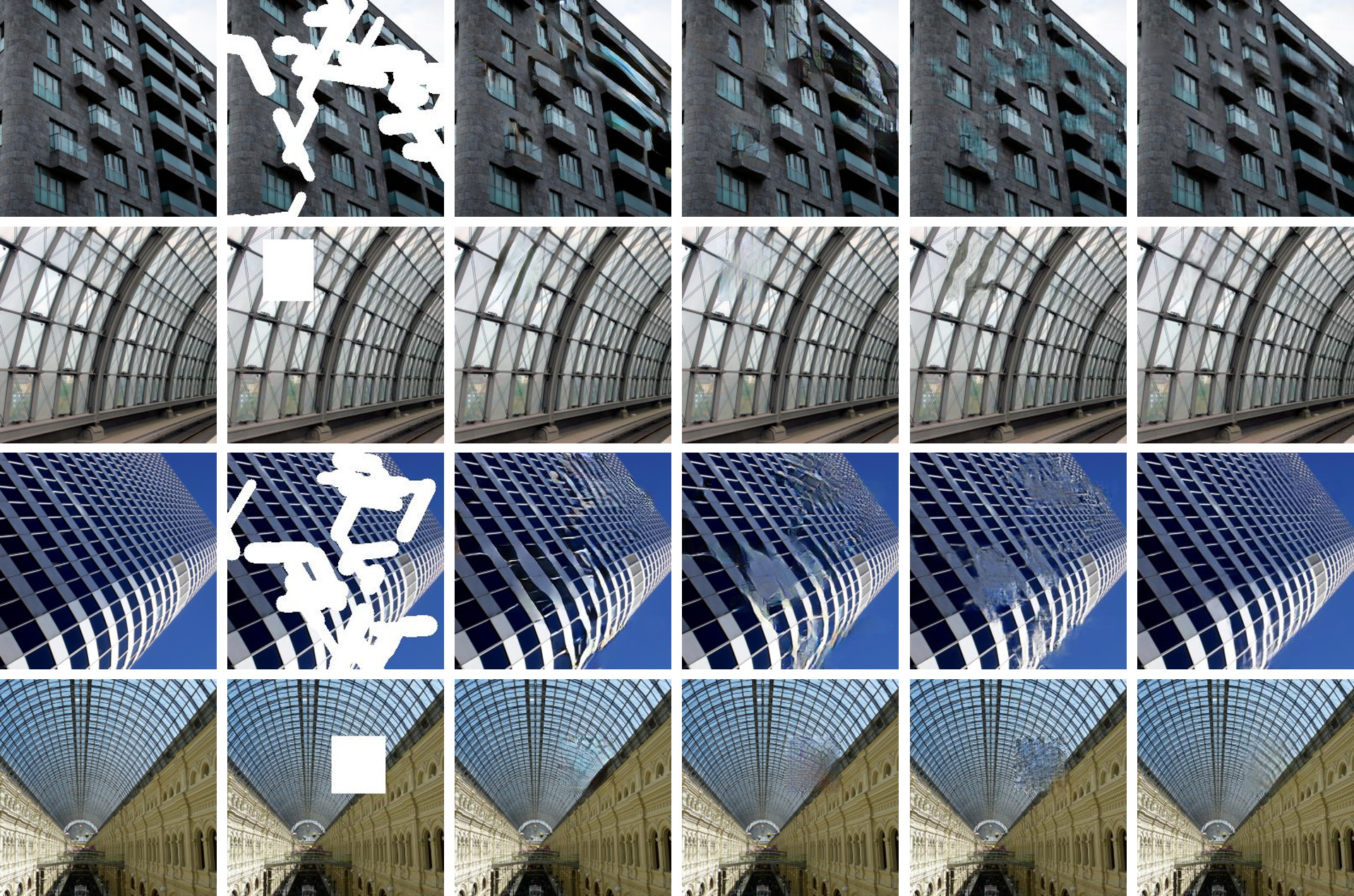}}
			\centerline{GT  \hspace{1.7cm}   Masked \hspace{1.7cm}   GC~\cite{GatedConv}  \hspace{1.7cm}  EC~\cite{EC} \hspace{1.7cm}  DIP~\cite{DIP} \hspace{1.7cm}   Ours}
		\end{minipage}
		\caption{Qualitative comparison results with Urban100~\cite{Urban100} dataset}
		\label{fig:Urban100-1}
	\end{figure}
	
	\newpage
	\begin{figure}[h]
		\begin{minipage}[b]{0.95\columnwidth}
			\centering
			\centerline{\includegraphics[width=1.0\textwidth]{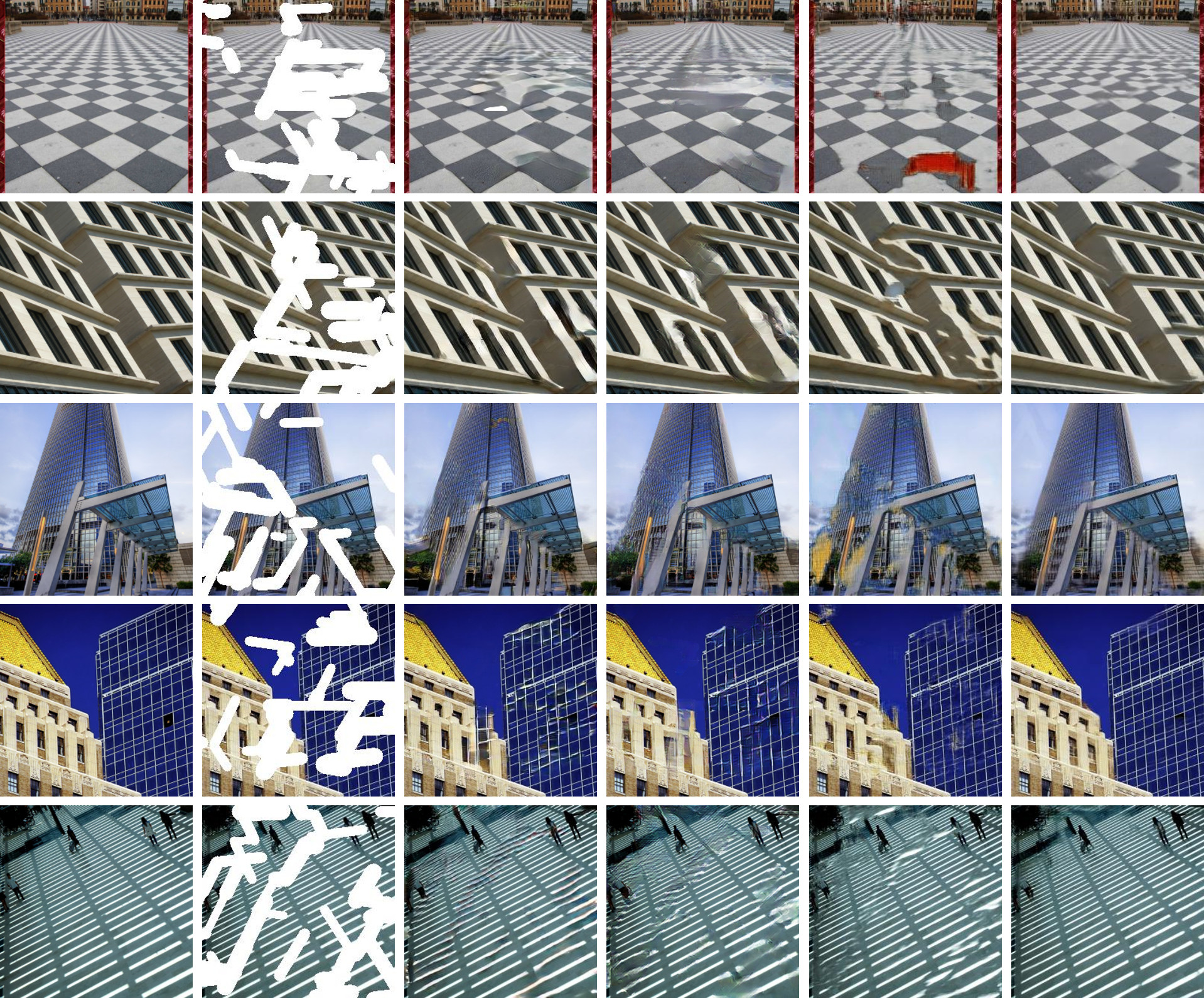}}
			\centerline{GT  \hspace{1.7cm}   Masked \hspace{1.7cm}   GC~\cite{GatedConv}  \hspace{1.7cm}  EC~\cite{EC} \hspace{1.7cm}  DIP~\cite{DIP} \hspace{1.7cm}   Ours}
		\end{minipage}
		\caption{Qualitative comparison results with Urban100~\cite{Urban100} dataset}
		\label{fig:Urban100-2}
	\end{figure}

	\subsubsection{Google Map~\cite{GoogleMap} Dataset}
	\begin{figure}[h!]
		\begin{minipage}[b]{0.95\columnwidth}
			\centering
			\centerline{\includegraphics[width=1.0\textwidth]{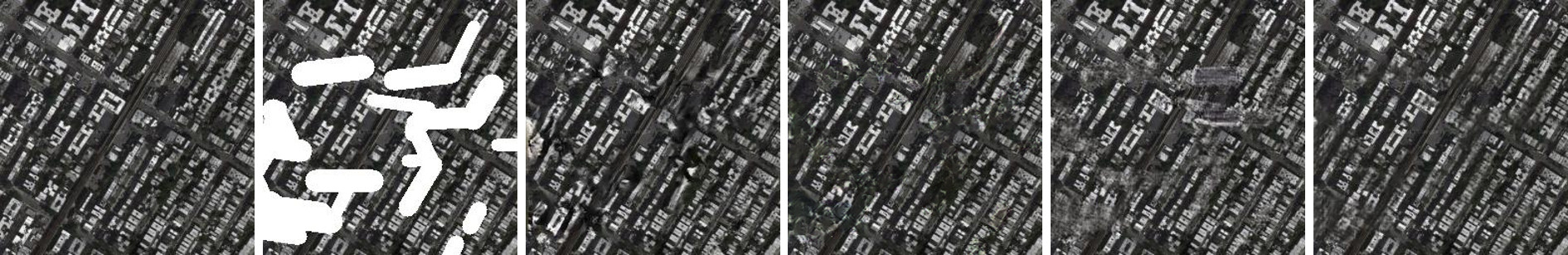}}
			\centerline{GT  \hspace{1.7cm}   Masked \hspace{1.7cm}   GC~\cite{GatedConv}  \hspace{1.7cm}  EC~\cite{EC} \hspace{1.7cm}  DIP~\cite{DIP} \hspace{1.7cm}   Ours}
		\end{minipage}
		\caption{Qualitative comparison results with Google Map~\cite{GoogleMap} dataset}
		\label{fig:GoogleMap-1}
	\end{figure}
	
	\newpage
	\begin{figure}[h]
		\begin{minipage}[b]{0.95\columnwidth}
			\centering
			\centerline{\includegraphics[width=1.0\textwidth]{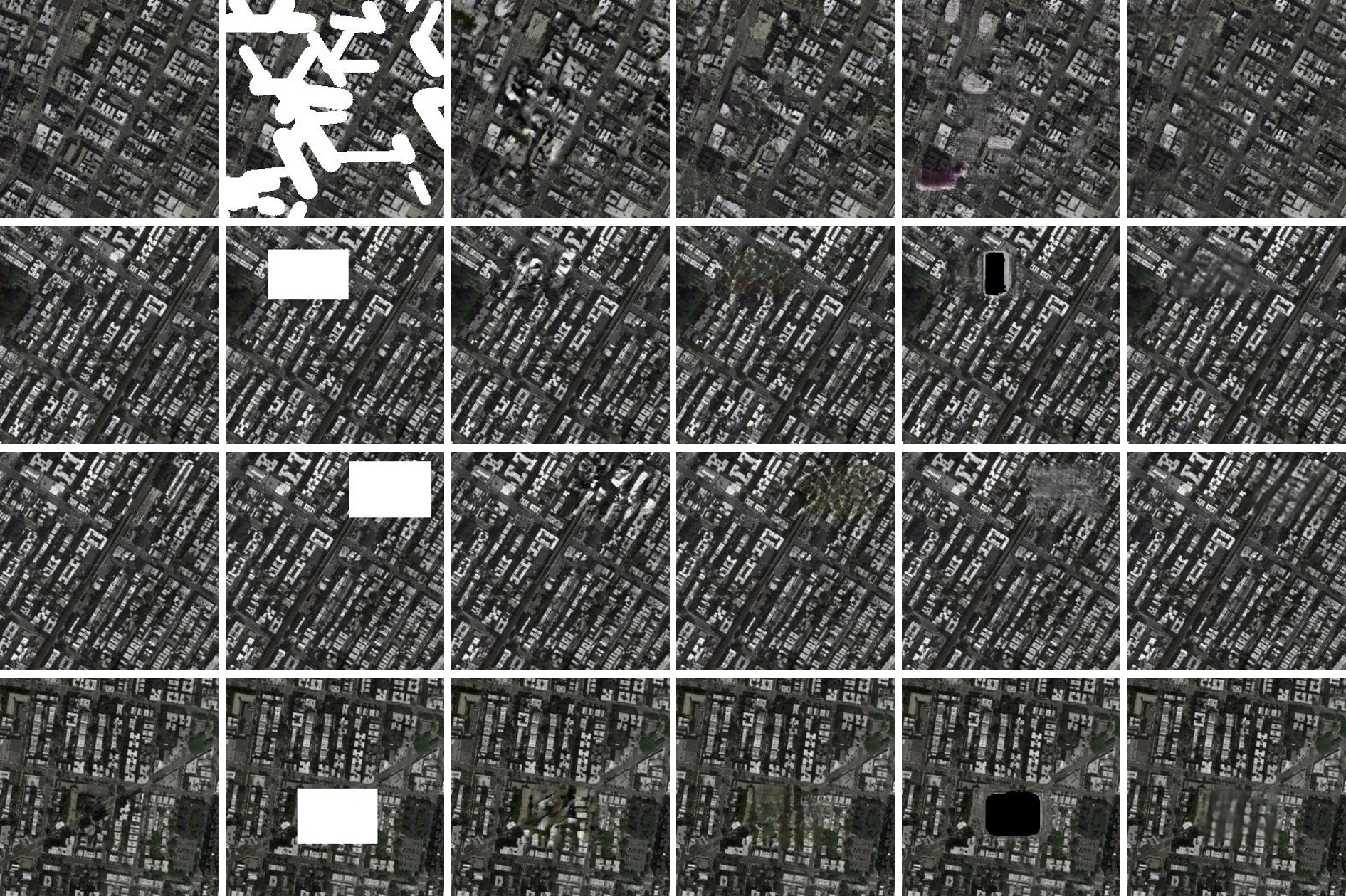}}
			\centerline{GT  \hspace{1.7cm}   Masked \hspace{1.7cm}   GC~\cite{GatedConv}  \hspace{1.7cm}  EC~\cite{EC} \hspace{1.7cm}  DIP~\cite{DIP} \hspace{1.7cm}   Ours}
		\end{minipage}
		\caption{Qualitative comparison results with Google Map~\cite{GoogleMap} dataset}
		\label{fig:GoogleMap-2}
	\end{figure}

	\subsubsection{Facade~\cite{Facade} Dataset}
	\begin{figure}[h!]
		\begin{minipage}[b]{0.95\columnwidth}
			\centering
			\centerline{\includegraphics[width=1.0\textwidth]{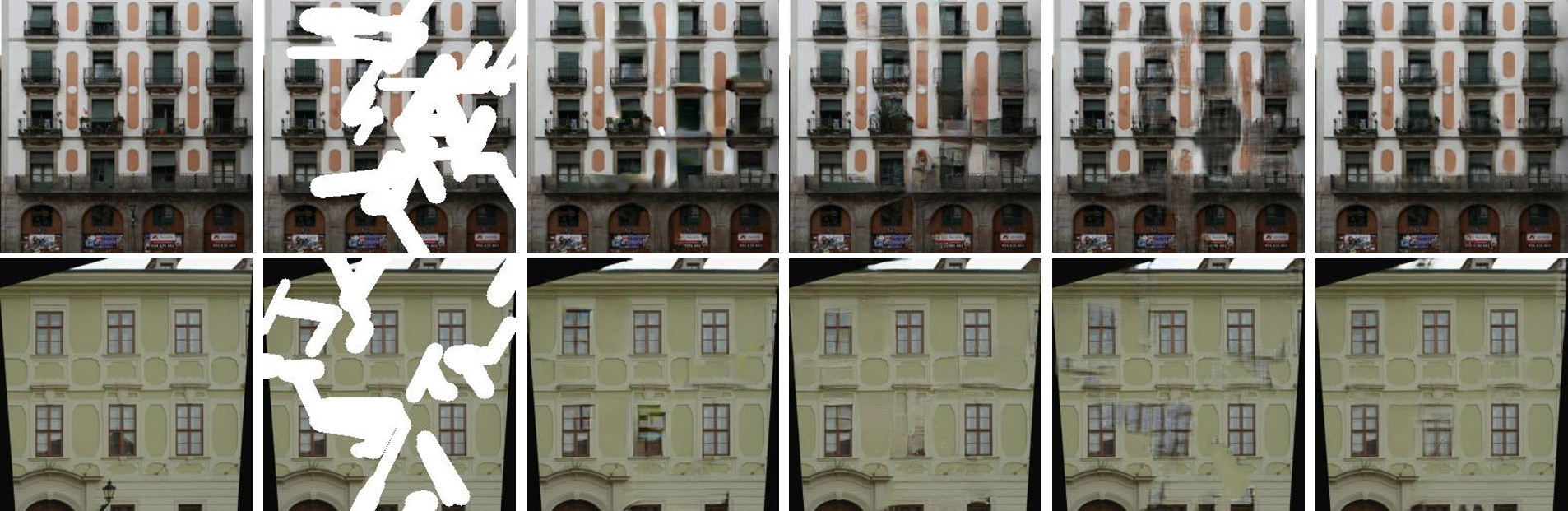}}
			\centerline{GT  \hspace{1.7cm}   Masked \hspace{1.7cm}   GC~\cite{GatedConv}  \hspace{1.7cm}  EC~\cite{EC} \hspace{1.7cm}  DIP~\cite{DIP} \hspace{1.7cm}   Ours}
		\end{minipage}
		\caption{Qualitative comparison results with Facade~\cite{Facade} dataset}
		\label{fig:Facade-1}
	\end{figure}
	
	\newpage
	\begin{figure}[h]
		\begin{minipage}[b]{0.95\columnwidth}
			\centering
			\centerline{\includegraphics[width=1.0\textwidth]{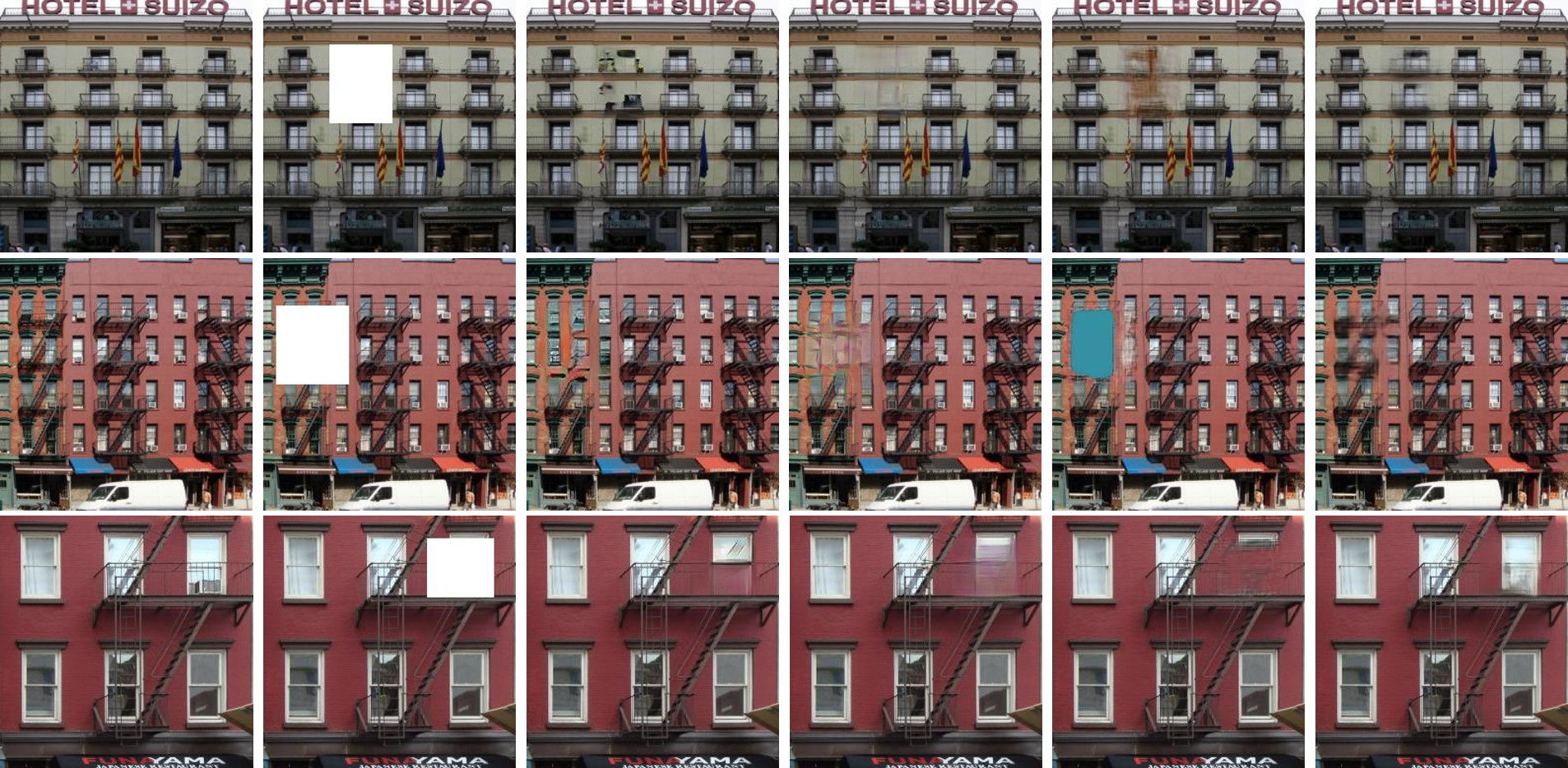}}
			\centerline{GT  \hspace{1.7cm}   Masked \hspace{1.7cm}   GC~\cite{GatedConv}  \hspace{1.7cm}  EC~\cite{EC} \hspace{1.7cm}  DIP~\cite{DIP} \hspace{1.7cm}   Ours}
		\end{minipage}
		\caption{Qualitative comparison results with Facade~\cite{Facade} dataset}
		\label{fig:Facade-2}
	\end{figure}

	\subsubsection{BCCD~\cite{BCCD} Dataset}
	\begin{figure}[h]
		\begin{minipage}[b]{0.95\columnwidth}
			\centering
			\centerline{\includegraphics[width=1.0\textwidth]{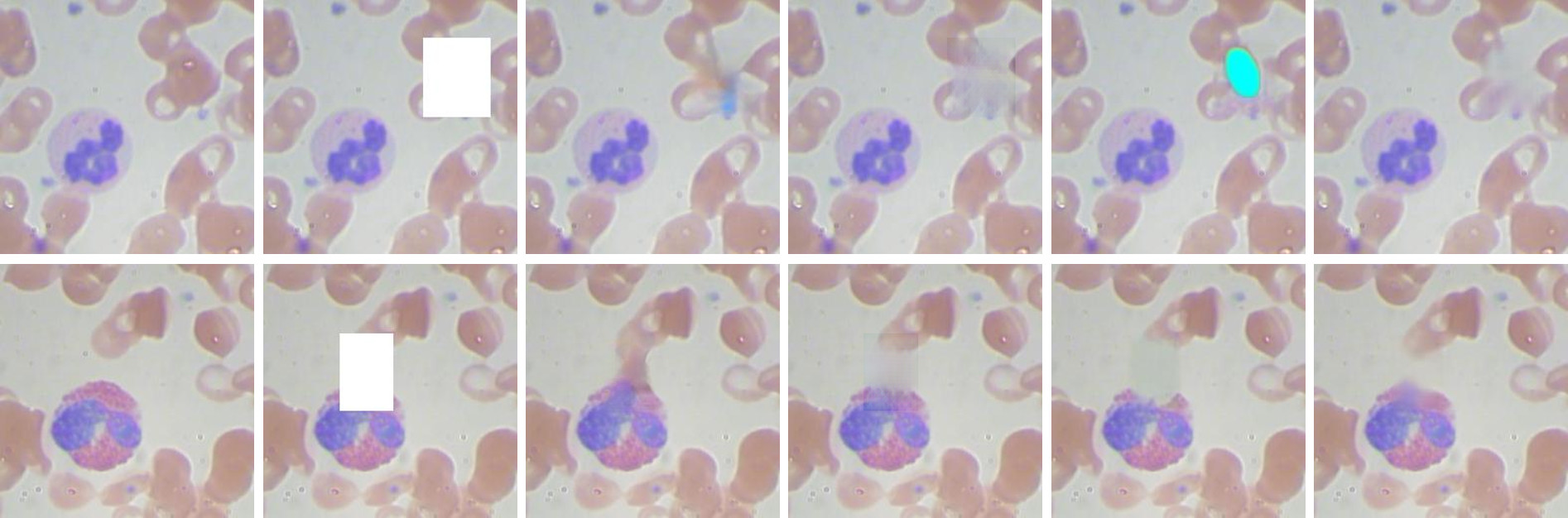}}
			\centerline{GT  \hspace{1.7cm}   Masked \hspace{1.7cm}   GC~\cite{GatedConv}  \hspace{1.7cm}  EC~\cite{EC} \hspace{1.7cm}  DIP~\cite{DIP} \hspace{1.7cm}   Ours}
		\end{minipage}
		\caption{Qualitative comparison results with BCCD~\cite{BCCD} dataset}
		\label{fig:BCCD}
	\end{figure}

	\newpage
	\subsubsection{KLH~\cite{KLH} Dataset}
	\begin{figure}[h]
		\begin{minipage}[b]{0.95\columnwidth}
			\centering
			\centerline{\includegraphics[width=1.0\textwidth]{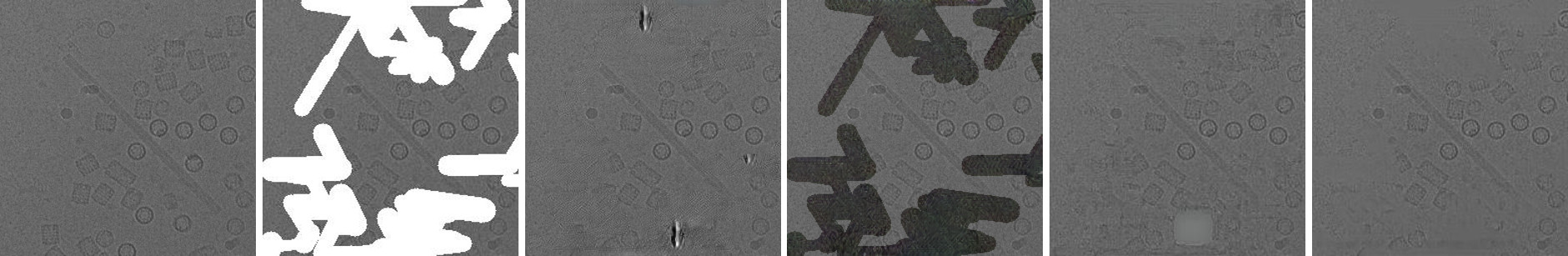}}
			\centerline{GT  \hspace{1.7cm}   Masked \hspace{1.7cm}   GC~\cite{GatedConv}  \hspace{1.7cm}  EC~\cite{EC} \hspace{1.7cm}  DIP~\cite{DIP} \hspace{1.7cm}   Ours}
		\end{minipage}
		\caption{Qualitative comparison results with KLH~\cite{KLH} dataset}
		\label{fig:KLH}
	\end{figure}

	\subsubsection{Document~\cite{Document} Dataset}
	\begin{figure}[h!]
		\begin{minipage}[b]{0.95\columnwidth}
			\centering
			\centerline{\includegraphics[width=1.0\textwidth]{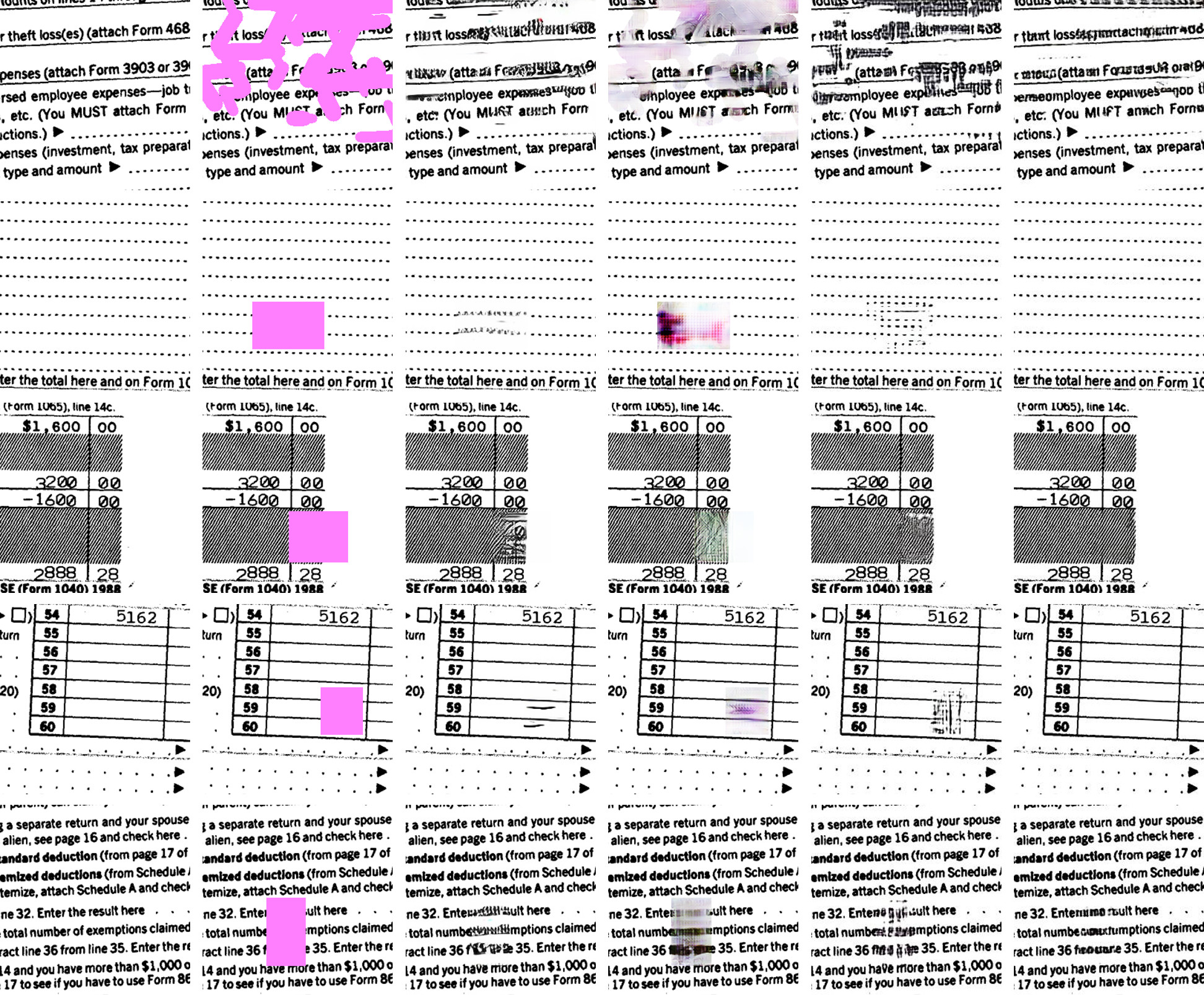}}
			\centerline{GT  \hspace{1.7cm}   Masked \hspace{1.7cm}   GC~\cite{GatedConv}  \hspace{1.7cm}  EC~\cite{EC} \hspace{1.7cm}  DIP~\cite{DIP} \hspace{1.7cm}   Ours}
		\end{minipage}
		\caption{Qualitative comparison results with Document~\cite{Document} dataset}
		\label{fig:Document}
	\end{figure}

	\vfill
	\pagebreak

	\bibliographystyle{IEEEbib}
	\bibliography{strings,refs}

\end{document}